\documentclass[journal]{IEEEtran}
\usepackage{amsmath,amsfonts}
\usepackage{algorithmic}
\usepackage{algorithm}
\usepackage{array}
\usepackage[utf8]{inputenc}

\usepackage[backend=bibtex,style=numeric,sorting=none]{biblatex}
\usepackage{booktabs}

\usepackage{graphicx}
\usepackage{hyperref}
\hypersetup{
    colorlinks=true,
    linkcolor=blue,
    filecolor=magenta,      
    urlcolor=cyan,
    pdftitle={Overleaf Example},
    pdfpagemode=FullScreen,
    }
\usepackage{subfig}

\usepackage{siunitx}
\usepackage{stfloats}
\usepackage{subcaption}
\usepackage{tabularx}
\usepackage{textcomp}
\usepackage{threeparttable}

\usepackage{url}

\usepackage{verbatim}

\begin{document}

\title{Lightweight Cloud Masking Models for On-Board Inference in Hyperspectral Imaging}

\author{Mazen Ali, Ant\'onio Pereira, Fabio Gentile, Aser Cortines, Sam Mugel, Rom\'an Or\'us, Stelios P. Neophytides, Michalis Mavrovouniotis \thanks{M. Ali, A. Pereira, F. Gentile, A. Cortines, S. Mugel, R. Or\'us are with Multiverse Computing, San Sebastian, Spain. email: \{mazen.ali, antonio.pereira, fabio.gentile, aser.cortines, sam.mugel, roman.orus\}@multiversecomputing.com} % 
\thanks{S. Neophtyides and M. Mavrovouniotis are with ERATOSTHENES Centre of Excellence and Cyprus University of Technology, Limassol, Cyprus. email: \{stelios.neophytides, michalis.mavrovouniotis\}@eratosthenes.org.cy}}



\maketitle

\begin{abstract}
Cloud and cloud shadow masking is a crucial preprocessing step in hyperspectral satellite imaging, enabling the extraction of high-quality, analysis-ready data. This study evaluates various machine learning approaches, including gradient boosting methods such as XGBoost and LightGBM as well as convolutional neural networks (CNNs). All boosting and CNN models achieved accuracies exceeding 93\%. Among the investigated models, the CNN with feature reduction emerged as the most efficient, offering a balance of high accuracy, low storage requirements, and rapid inference times on both CPUs and GPUs. Variations of this version, with only up to 597 trainable parameters, demonstrated the best trade-off in terms of deployment feasibility, accuracy, and computational efficiency. These results demonstrate the potential of lightweight artificial intelligence (AI) models for real-time hyperspectral image processing, supporting the development of on-board satellite AI systems for space-based applications.

\end{abstract}


\IEEEpeerreviewmaketitle

\section{Introduction}
\IEEEPARstart{S}{atellite} remote sensing is essential for Earth observation, supporting downstream applications in environmental monitoring, disaster response, and land cover classification \cite{10669817}. However, optical satellite imaging is often obstructed by clouds, necessitating cloud masking to retain only high-quality, unobstructed data for analysis \cite{app14072887}. 

Cloud masking methods generally fall into three categories:  
\begin{itemize}
    \item Physics-based methods use spectral thresholds to detect clouds efficiently but struggle with complex cloud structures \cite{sen2cor, fmask}.  

    \item Machine learning methods classify pixels based on spectral patterns but require labeled training data and adaptation for different sensors \cite{s2cloudless}.  

    \item Deep learning methods leverage CNNs to capture spatial and spectral patterns, achieving high accuracy but demanding significant computational resources \cite{clouds2mask, justo24}.
\end{itemize}

Hyperspectral Imaging (HSI) extends traditional remote sensing by capturing detailed spectral information across numerous bands, enabling precise surface and atmospheric characterization \cite{8113122}. However, the high dimensionality of HSI data makes cloud and cloud shadow detection challenging, as clouds exhibit diverse spectral signatures that can overlap with bright surfaces like snow, ice, and sand \cite{ZHAI2018235}. Traditionally, AI-based cloud masking has been performed on the ground due to its computational cost. However, advancements in onboard processing, including lightweight deep learning models and specialized hardware (e.g., FPGAs, VPUs), now enable AI-driven cloud segmentation directly on satellites.

As the era of CubeSats arrived with various on-board processing implementations, low energy consumption models have to be developed, due to satellites constraints \cite{shakoor_comprehensive_2025}. Typically, a CubeSat of single, double or triple unit, operates with body mounted solar panels, generating limited power of about 5-8 watts \cite{es1315}. At the same time, payloads such as integrated Field-Programmable Gate Arrays and Vision Processing Units, employed for the on-board processing tasks, in hyperspectral images, often require an energy consumption of about \qty{12}{\watt} \cite{6187240,Lub2024}.

This study focuses on leveraging novel techniques to improve model size reduction for higher utilization of on-board hardware and provision of analysis ready data directly from satellites. Our aim is to develop lightweight models for satellite on-board cloud masking in hyperspectral images.
Specifically, 1) develop a high accuracy cloud detector using a public dataset, 2) reduce model sizes of CNNs, 3) test classical machine learning models as lightweight AI model alternatives for on-board processing and
4) integrate metadata-driven detection to optimize data utilization and reduce unnecessary down-linking.

The paper is structured as follows. Section \ref{sec:methods} outlines the investigated models. Section \ref{sec:results} presents performance evaluations. Section \ref{sec:cots_compatibility} discusses onboard processing feasibility. We provide a brief summary of the results in Section \ref{sec:conclusions}.

\section{Related Works}

The first ever on-board AI demonstrator was a cloud detector on $\Phi$-Sat 1 satellite \cite{9600851}. Upon the success of the
$\Phi$-Sat missions by the European Space Agency, the benefit from other various satellite on-board applications (e.g., vessel detection, image compression, crop health, wildfires, etc) has been identified \cite{marin_-sat-2_2021, 10623617}. For example, on-board image compression can open new pathways for intelligent data processing as it reduces the needs for transmission bandwidth and storage \cite{10185543}. In addition, solutions like CloudScout, an on-board cloud detector for hyperspectral imagery, can reduce the amount of data transmitted to satellite data acquisition stations by discarding cloudy pixels with a cloud detection accuracy of 92\% \cite{rs12142205}. An innovative space-grade Field-Programmable Gate Array (FPGA) was designed to host CloudScout pre-trained model with a 45\% reduction in memory footprint with insignificant loss in accuracy, 24\% reduction in energy consumption and a 2.4x speed-up at the cost
of 1.8x higher power consumption in a benchmark comparison to the well-known Myriad 2 Vision Processing Unit \cite{rs13081518}.

An enhanced variation of the
\textit{function of mask (Fmask)} algorithm for segmentation of clouds on hyperspectral images derived by Gaofen mission showed a
high accuracy of \qty{95.51}{\percent} \cite{rs13234936}. Moreover, a dual-branch CNN for Gaofen-5 reached an F1-score of \qty{94}{\percent} \cite{rs12132106}. A cloud detection model using hyperspectral layers related to atmospheric column water vapor from PRecursore IperSpettrale della Missione Applicativa and Airborne Visible InfraRed Imaging Spectrometer-Next Generation satellite missions achieved an overall accuracy ranging from \qtyrange{97}{100}{\percent} on both sensors \cite{10443884}. In addition, transfer learning techniques in fundamental NN architectures like ResNet50 and VGG16 showed that with a small dataset of 972 samples an accuracy over
\qty{90}{\percent} can be reached for cloud detection in hyperspectral images \cite{10.3389/fenvs.2022.1039249}. Despite those studies showing
a great capability in capturing the complex patterns of cloud detection, none of them investigated the potential of on-board processing.

More research studies like those for Hyper-Spectral Small Satellite for Ocean Observation \cite{justo24,Justo25} are needed in the field of satellite on-board cloud detection with a special focus on the hardware constraints of processing units and the satellite itself. Thus, in this research study we are investigating new ways for reducing model size without significant accuracy trade-offs.

\section{Methods}\label{sec:methods}
Evaluating the task, type of data and available literature, we have selected the
following approaches to investigate:
\texttt{XGBoost}, \texttt{LightGBM} and CNNs.
In the following, we give a brief description of each method.

\subsection{\texttt{XGBoost}}
Extreme Gradient Boosting (\texttt{XGBoost} \cite{xgboost}) is an ensemble learning method that has demonstrated strong performance across a wide range of applications \cite{xgboostwins},
including cloud masking \cite{s2cloudless}. In ensemble learning, a robust predictive model is constructed by combining multiple weaker models.

In \texttt{XGBoost}, the weak learners are, by default, regression trees, also known as classification and regression trees (CARTs). The model iteratively enhances its predictions by adding new regression trees to the ensemble. \texttt{XGBoost} employs a level-wise tree growth strategy, meaning that at each growth step, the algorithm refines predictions by successively splitting each leaf node into two. As a result, the weak learners in the ensemble are balanced trees with $2^{\mathrm{max\_depth}}$ terminal nodes.

This approach is well-suited for handling large datasets with high-dimensional feature spaces, making it particularly relevant to our study. Additionally, the \texttt{XGBoost} software package \cite{xgboostsoft} provides support for distributed learning and efficient large-scale data processing.

\subsection{\texttt{LightGBM}}
A potential limitation of \texttt{XGBoost} is that the number of leaves increases exponentially with tree depth, which can lead to computational inefficiencies.
To mitigate this issue, particularly in scenarios requiring deeper trees, \texttt{LightGBM} offers an alternative approach.

Unlike \texttt{XGBoost}, which expands trees in a level-wise manner, \texttt{LightGBM} employs a leaf-wise growth strategy. It selectively expands only the leaf nodes where splitting a feature yields the greatest reduction in residual error. Consequently, the weak learners in \texttt{LightGBM} form unbalanced trees that can reach greater depths while maintaining a significantly lower number of leaves compared to \texttt{XGBoost}.

These optimizations, along with additional enhancements in the \texttt{LightGBM} library \cite{lgbmsoft}, enable more efficient training on large-scale datasets. However, a potential drawback of this approach is its susceptibility to overfitting, particularly when tree depth is excessively high.

\subsection{Convolutional Neural Networks}
Convolutional neural networks (CNNs) are among the most widely used architectures for image segmentation. The richness of spectral features in the
dataset \cite{hypso1dataset}
allows for the use of relatively small CNN models while still achieving high predictive accuracy \cite{justo24}.

In \cite{justo24}, the authors obtained optimal results using a compact 1D CNN model. Here,
``1D'' refers to convolutions being applied exclusively along the spectral dimension for each pixel independently, without incorporating spatial context from neighboring pixels. As a baseline benchmark, we implement the same 1D CNN model in \texttt{PyTorch}, following the architecture described in \cite{justo24}.
However, we introduce a different data normalization technique -- Z-score normalization -- rather than the original method, which leads to an improvement in test accuracy.

To further reduce model size, we use Singularity\texttrademark\ to apply model compression based on the singular value decomposition of the convolutional layers in the CNN model \cite{praggastis2022}.

Finally, we observed that on CPUs inference speed was primarily constrained by the high dimensionality of the input features. To address this, we used the principal component analysis tools (PCA) integrated within Multiverse Computing’s Singularity\texttrademark\ suite to reduce the feature space from 112 to 30, 18, 7 and 4 dimensions.
We refer to these models as \texttt{1DJuLiNet}, \texttt{1DJuLiNetSingularity}, and \texttt{1DJuLiNetSingularityFx}, with $\texttt{x}\in\{30, 18, 7, 4\}$,
respectively.

\section{Experimental Results}\label{sec:results}

\subsection{Experimental Setup} 
All experiments are performed on a \texttt{g5.2xlarge} AWS instance, equipped with an AMD EPYC 7R32 processor and a single NVIDIA A10G Tensor Core GPU with 24 GB of memory. The AMD EPYC 7R32 is a server-grade processor based on the Zen 2 architecture, featuring 8 physical cores, 16 threads (limited to 8 by the \texttt{g5.2xlarge} instance), and a base clock speed of 3.2 GHz. The A10G GPU includes third-generation NVIDIA Tensor Cores optimized for machine learning workloads, delivering up to 250 TOPS of performance and 24 GB of dedicated memory. Full hardware specifications can be found on the AWS website.

A basic hyperparameter tuning was conducted for the boosting models while the hyperparameters for the CNN based models were kept identical to \cite{justo24}.
The reported experimental results correspond to the final optimized models only. For \texttt{XGBoost}, we performed a three-stage hyperparameter search using optuna (\url{https://optuna.org/} \cite{optuna})
with limits and optimal parameters listed in Table \ref{tab:xgboost_params}.
While for \texttt{LightGBM} a single stage hyperparameter search was performed using the same tooling, see Table \ref{tab:lightgbm_params}. 

\begin{table}[!t]

    \centering
    \caption{Table representing the search space and optimal parameters for the \texttt{XGBoost} model.}
    \label{tab:xgboost_params}
    \begin{tabular}{lllll}
    \toprule
        Stage    &  Parameter Name               & Type     & Range         & Optimal\\
    \midrule
        1        & \texttt{max\_depth}           & \texttt{int    }  & 3 -- 12       & 5      \\
        1        & \texttt{min\_child\_weight}   & \texttt{int    }  & 1 -- 10       & 3      \\
        2        & \texttt{subsample}            & \texttt{float32}  & 0.5 -- 1.0    & 0.504  \\
        2        & \texttt{colsample\_bytree}    & \texttt{float32}  & 0.5 -- 1.0    & 0.779  \\
        3        & \texttt{learning\_rate}       & \texttt{float32}  & 0.01 -- 0.5   & 0.258  \\
        3        & \texttt{num\_boost\_round}    & \texttt{int    }  & 50 -- 1000    & 695    \\
    \bottomrule
    \end{tabular}
    
\end{table}

\begin{table}[!t]

    \centering
    \caption{Table representing the search space and optimal parameters for the \texttt{LightGBM} model.}
    \label{tab:lightgbm_params}
    \begin{tabular}{lllll}
    \toprule
        Stage    &  Parameter Name                & Type               & Range            & Optimal\\
    \midrule
        1        & \texttt{num\_leaves}           & \texttt{int    }   & 20 -- 1000       & 28      \\
        1        & \texttt{min\_data\_in\_leaf}   & \texttt{int    }   & 20 -- 2000       & 410      \\
        1        & \texttt{num\_boost\_round}     & \texttt{int    }   & 50 -- 200        & 140    \\
    \bottomrule
    \end{tabular}
    
\end{table}
\subsection{Dataset Description}
To develop and evaluate our cloud masking models, a labeled hyperspectral dataset of satellite imagery is required.
Such data is scarce due to the high cost and complexity of hyperspectral imaging missions. The authors in \cite{hypso1dataset} prepared one such
suitable dataset, which is obtained from the HYPSO-1 mission and it is utilized in our experiments. The dataset provides labeled hyperspectral images captured by a 6U CubeSat carrying a hyperspectral imager.

The HYPSO-1 sensor system consists of a spectrometer with a 684-pixel slit, which sequentially scans 956 line frames to form a complete hyperspectral image, see \cite{hypso1mission, hypso1design}. The resulting images have a spatial resolution of approximately \qty{100}{\meter} $\times$ \qty{600}{\meter}, covering a wide area with high spectral fidelity. Each image comprises 120 spectral bands, spanning the \qty{400}{\nano\meter} to \qty{800}{\nano\meter} range with a spectral resolution of approximately \qty{5}{\nano\meter}. This configuration enables detailed spectral analysis of surface and atmospheric conditions.  

To ensure data consistency and accuracy, the raw sensor measurements undergo a two-step calibration process \cite{hypso1dataset}. Spectral calibration is performed using a second-order polynomial fit, assigning precise wavelengths to each spectral band and storing them as metadata. Radiometric calibration then converts the raw digital number values into physical radiance values. Note that, for our study,
only the radiance-calibrated data is utilized, as prior work has demonstrated its superior performance for hyperspectral analysis \cite{hypso1dataset, justo24}.  

The HYPSO-2 Sea-Land-Cloud-Labeled dataset consists of 38 labeled images, following the same train-validation-test split as the original study \cite{justo24}. Specifically, 30 images are used for training, 3 images for validation, and 5 images for testing. In terms of pixels, the total is around
20, 2, and 3 million pixels for training, validation and testing, respectively, where each pixel consists of 120 spectral features. Note that 8 spectral bands are excluded due to known anomalies related to radiometric calibration or atmospheric light absorption as described in the original study. This leaves us with 112 spectral channels per pixel for model training, validation and testing. The dataset is openly accessible for download at the following website: \url{https://ntnu-smallsat-lab.github.io/
 hypso1_sea_land_clouds_dataset/.}

\begin{table*}[!t]
    \centering
    \caption{Comparison of model size and training time for various cloud masking models. All models, except \texttt{LightGBM}, are trained on GPUs.}
    \label{tab:train-time}
    \begin{threeparttable}
    \begin{tabular}{@{} lccc @{}}
        \toprule
        \textbf{Model} & \textbf{Size (Parameter $\#$)} &\textbf{Size (\unit{\kilo\byte})} & \textbf{Training Time (\unit{\minute})} \\
        \midrule
        \texttt{XGBoost}                  & \textbf{66} trees, 4104 nodes                      & 264 (\texttt{float64}) & 13 \\
        \texttt{LightGBM}                 & 69 trees, \textbf{1932} nodes                      & \textbf{228} (\texttt{float64}) & \textbf{12}  \\
        \midrule
        \texttt{1DJuLiNetRetrained}            & 4563 & 24 (\texttt{float32}) & 42 \\
        \texttt{1DJuLiNetSingularity}  & 1419 & 12 (\texttt{float32}) & 50 \\
        \texttt{1DJuLiNetSingularityF30}         & 597 trainable, 4554 total                    & 20 (\texttt{float32}) & 31 \\
        \texttt{1DJuLiNetSingularityF18}         & 525 trainable, 2541 total                    & 13 (\texttt{float32}) & 30 \\
        \texttt{1DJuLiNetSingularityF07}         &  63 trainable,  847 total                    &  6 (\texttt{float32}) & 28 \\
        \texttt{1DJuLiNetSingularityF04}         & \textbf{ 12} trainable,  \textbf{487} total                    &  \textbf{5} (\texttt{float32}) & \textbf{27} \\
        \bottomrule
    \end{tabular}
    \vspace{1em}
    \end{threeparttable}
\end{table*}

%%%%%%%%Figure 1%%%%%%%%%%
\begin{figure}[!t]
    \centering    
    \includegraphics[width=0.95\linewidth]{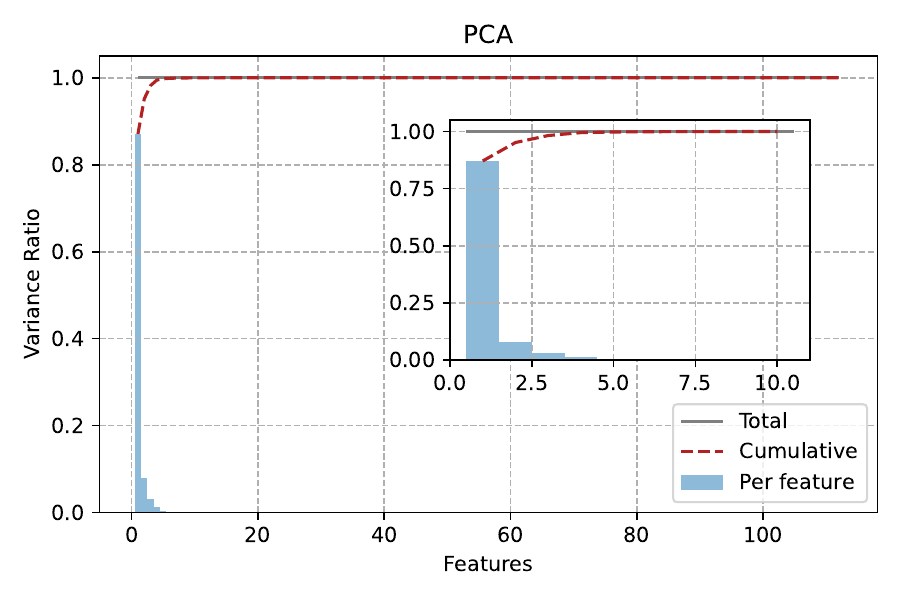}
    \caption{Histogram of the principal components in decreasing order of amplitude and their cumulative distribution. The inset shows a zoom-in of the first 10 components.}
    \label{fig:hist}
\end{figure}

\subsection{Model Training Comparison Experiment}
While training time is not intended as a strict performance metric
for on-board inference, it serves as an important indicator of computational demands and practical usability for preparing such models.
All CNN models are trained for two epochs, following the approach in \cite{justo24}, as additional epochs did not yield improvements in validation accuracy.

Note that in the feature-reduced CNN models (\texttt{1DJuLiNetSingularityFx}), the non-trainable parameters correspond to the projection matrix in the initial layer, which projects the original feature space to a lower-dimensional vector space. To produce feature-reduced models, the most prominent components are plotted in decreasing order in Fig.~\ref{fig:hist}. 
We observe that the first 7 main components account for over
\qty{99.9}{\percent} of the total variance,
while the largest 30 account for over \qty{99.99}{\percent} of the total variance.
Therefore, four different feature-reduced CNN model variants are investigated
-- \texttt{1DJuLiNetSingularityFx}, with $\texttt{x}\in\{30, 18, 7, 4\}$,
i.e.\, 30, 18, 7 and 4 features, respectively.

Table \ref{tab:train-time} shows the results of the trained model sizes and an estimate of the training time. The results indicate that boosting models,
i.e., \texttt{XGBoost} and \texttt{LightGBM} are the fastest to train while maintaining relatively small parameter counts and model sizes. Among the CNN-based models, the compressed CNN (\texttt{1DJuLiNetSingularity}) together with \texttt{1DJuLiNetSingularityF07} and \texttt{1DJuLiNetSingularityF04} have the smallest overall parameter count and model size. It is worth mentioning that
the two CNNs with feature reduction have the lowest number of trainable parameters, allowing them to achieve training speeds comparable to the booster models.

\begin{table*}[!t]
    \centering
    \caption{Comparison of model accuracy metrics.
   % All experiments were performed on a \texttt{g5.2xlarge} instance; hardware specifications can be found on AWS.
    All metrics given in \unit{\percent} and they are computed on the test set for the cloud class only (cloud/no cloud).
    }
    \label{tab:accuracy}
    \begin{threeparttable}
    \begin{tabular}{@{} lcccccc @{}}
        \toprule
        \textbf{Model} & \textbf{Accuracy} &\textbf{BOA} & \textbf{Precision} & \textbf{Recall} & \textbf{F1} & \textbf{Jaccard} \\
        \midrule
        \texttt{XGBoost}  & 93.32 & 90 & 92.7 & 78.32 & 84.91 & 73.78 \\
        \texttt{LightGBM} & 93.33 & 90.04 & 89.33 & 78.59 & 83.62 & 71.85  \\
        \texttt{1DJuLiNetRetrained} & \textbf{95.38} & \textbf{93.77} & 93.04 & \textbf{88.18} & \textbf{90.54} & \textbf{82.72} \\
        \texttt{1DJuLiNetSingularity} & 93.47 & 90.91 & 93.8 & 82.27 & 87.67 & 78.04 \\
        \texttt{1DJuLiNetSingularityF30} & 94.48 & 92.85 & 93.15 & 86.63 & 89.77 & 81.44 \\
        \texttt{1DJuLiNetSingularityF18} & 91.60 & 90.20 & 81.59 & 84.21 & 82.88 & 70.77 \\
        \texttt{1DJuLiNetSingularityF07} & 94.00 & 91.05 & \textbf{96.51} & 80.78 & 87.94 & 78.48 \\
        \texttt{1DJuLiNetSingularityF04} & 92.93 & 88.11 & 94.47 & 71.39 & 81.33 & 68.53 \\
        \bottomrule
    \end{tabular}
    \vspace{1em}
    \end{threeparttable}
\end{table*}

\begin{figure*}[!t]
     \subfloat[\texttt{XGBoost}]{\includegraphics[scale=0.4]{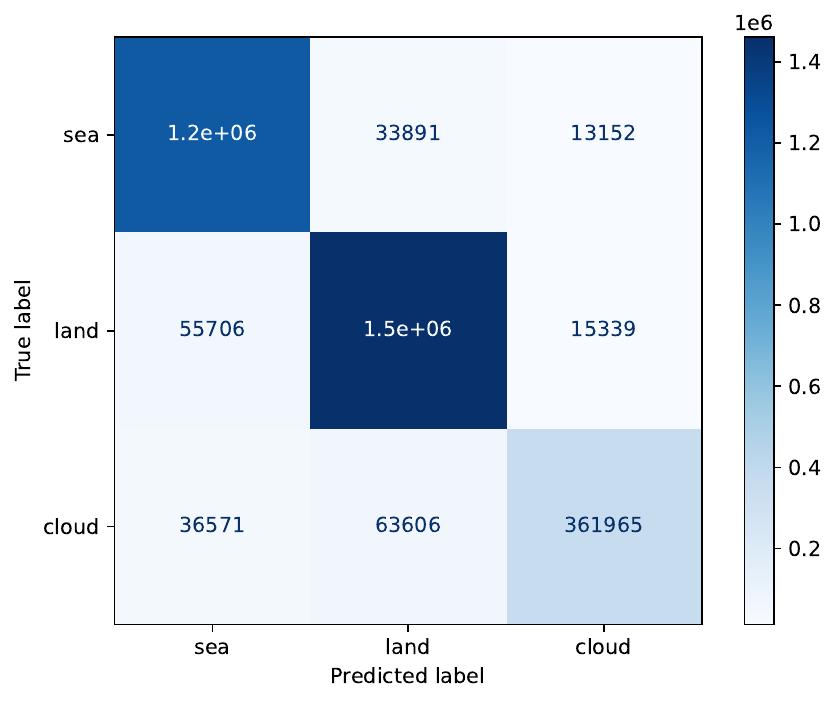}}
    \subfloat[\texttt{LightGBM}]{\includegraphics[scale=0.4]{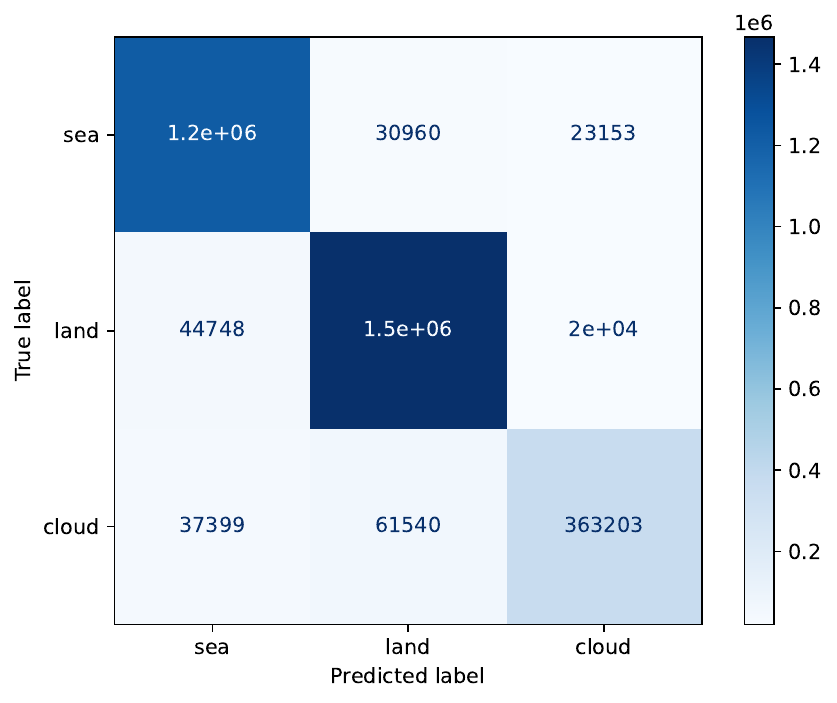}} 
    \subfloat[\texttt{1DJuLiNet}]{\includegraphics[scale=0.4]{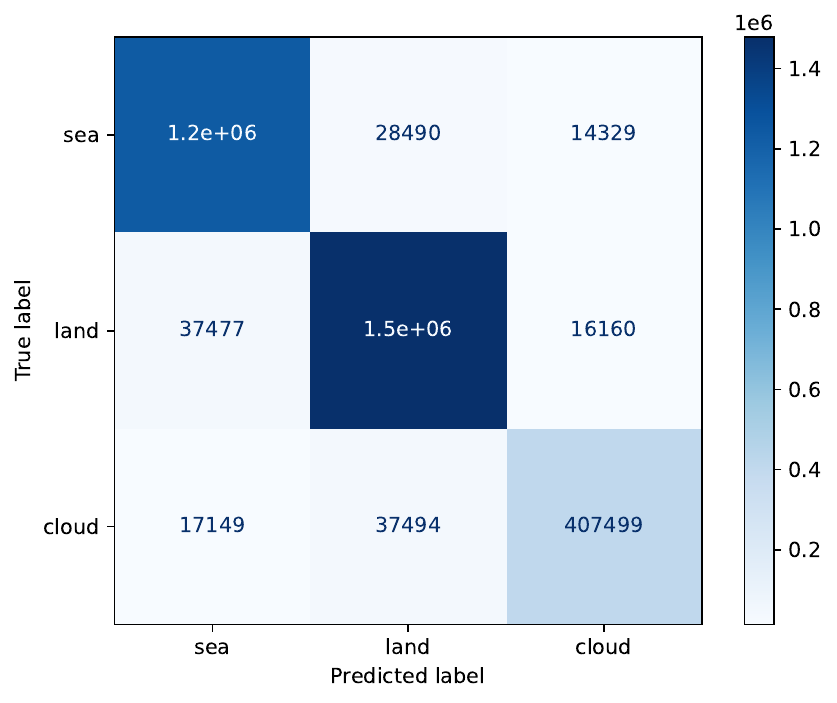}} \\
    \subfloat[\texttt{1DJuLiNetSingularity}]{\includegraphics[scale=0.4]{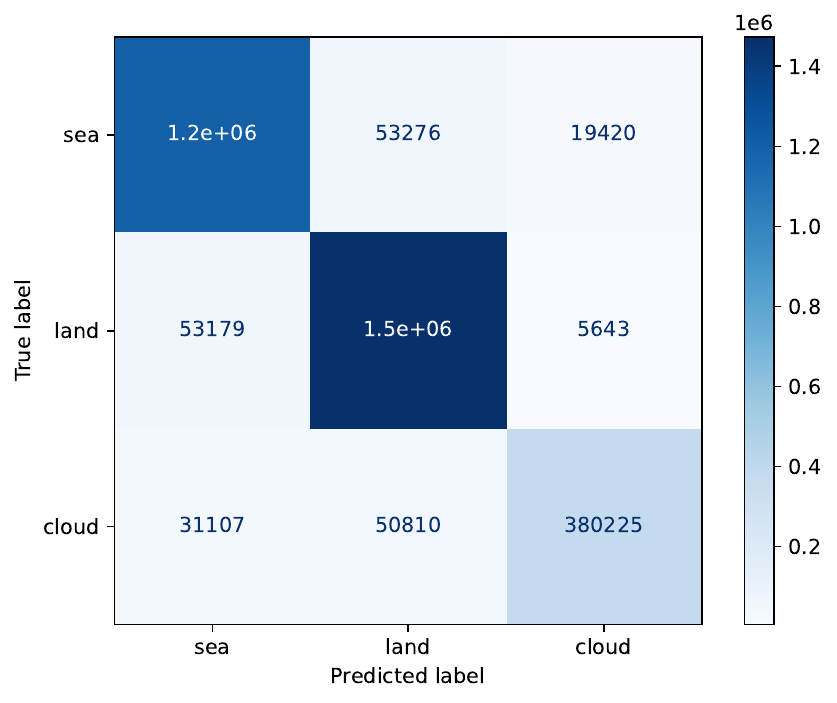}} 
    \subfloat[\texttt{1DJuLiNetSingularityF04}]{\includegraphics[scale=0.4]{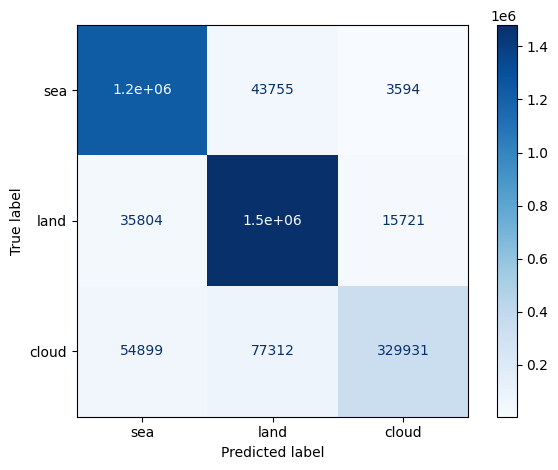}}
    \subfloat[\texttt{1DJuLiNetSingularityF07}]{\includegraphics[scale=0.4]{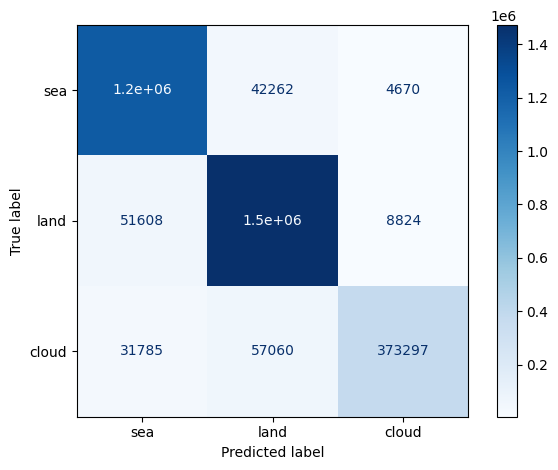}} \\
    \subfloat[\texttt{1DJuLiNetSingularityF18}]{\includegraphics[scale=0.4]{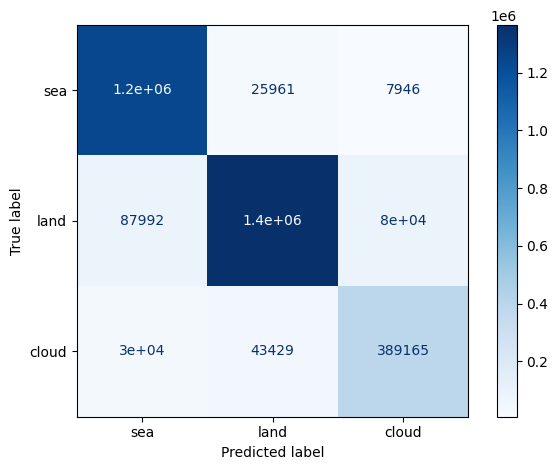}}
    \subfloat[\texttt{1DJuLiNetSingularityF30}]{\includegraphics[scale=0.4]{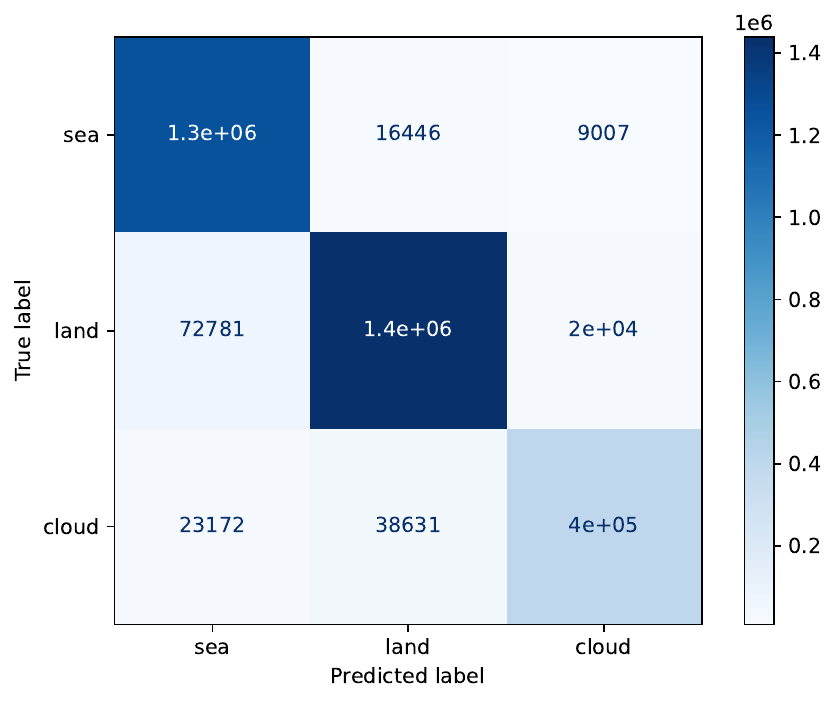}}
    \caption{Confusion matrices for different models.}
    \label{fig:confusion}
\end{figure*}

\begin{figure*}[!t]
    \centering
\subfloat[Ground Truth]{\includegraphics[scale=0.42]{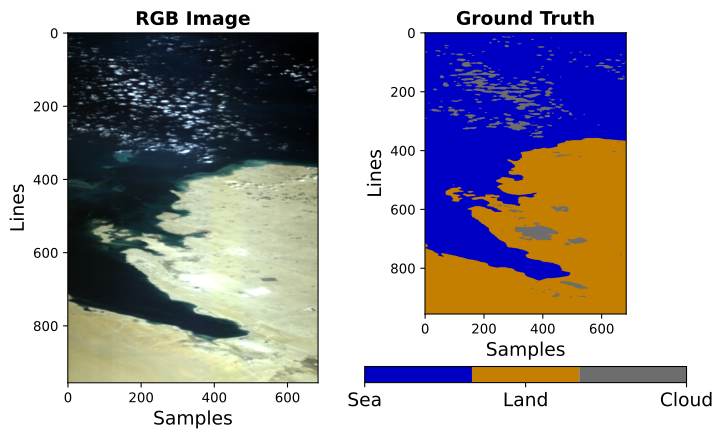}} \hspace{4mm}
\subfloat[\texttt{XGBoost}]{\includegraphics[scale=0.42]{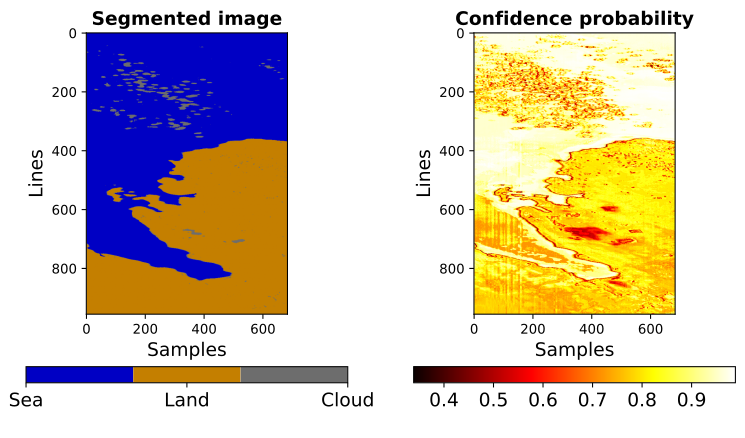}}\\
\subfloat[\texttt{1DJuLiNet}]{\includegraphics[scale=0.42]{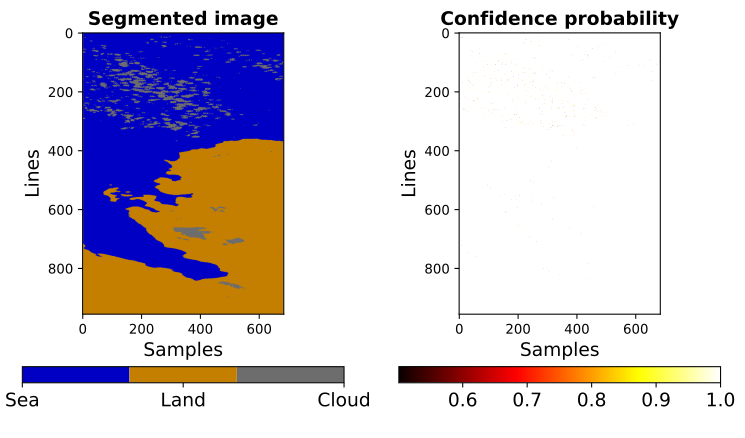}} \hspace{4mm}
\subfloat[\texttt{1DJuLiNetSingularity}]{\includegraphics[scale=0.42]{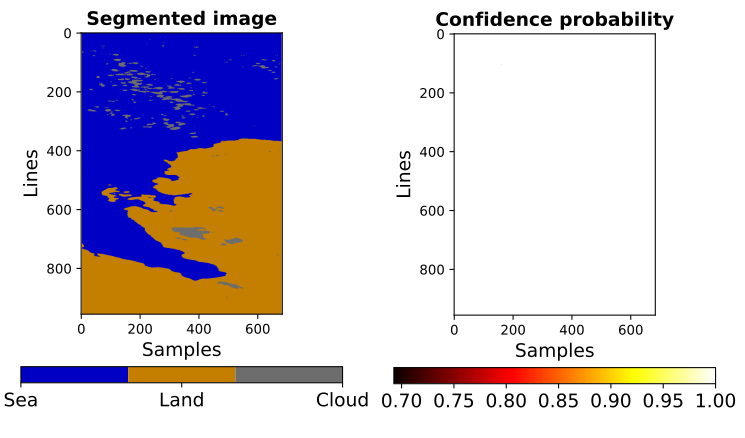}} \\
\subfloat[\texttt{1DJuLiNetSingularityF07}] {\includegraphics[scale=0.42]{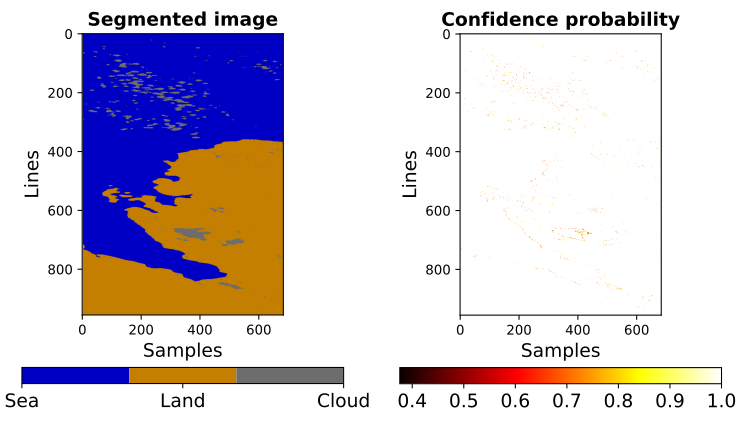}} \hspace{4mm}
\subfloat[\texttt{1DJuLiNetSingularityF30}] {\includegraphics[scale=0.42]{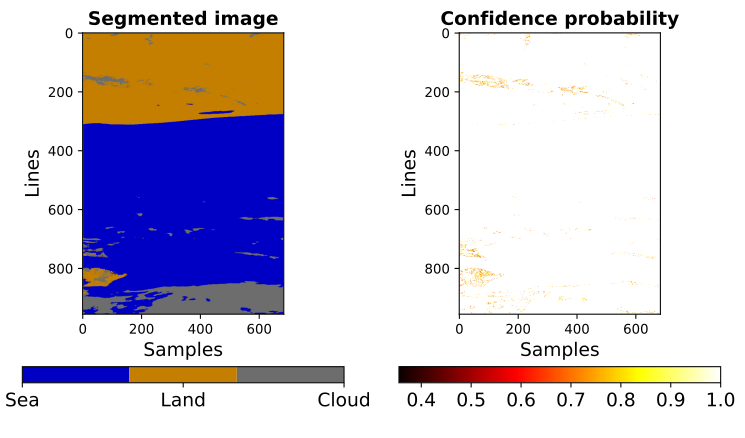}} 

\caption{Comparison of (a) Ground truth, (b) \texttt{XGBoost}, (c) \texttt{1DJuLiNet}, (d) \texttt{1DJuLiNetSingularity}, (e) \texttt{1DJuLiNetSingularityF07}, and (f) \texttt{1DJuLiNetSingularityF30} on the sample representing Qatar.}
\label{fig:sample1}
\end{figure*}

\begin{figure*}[!t]
    \centering
\subfloat[Ground Truth]{\includegraphics[scale=0.42]{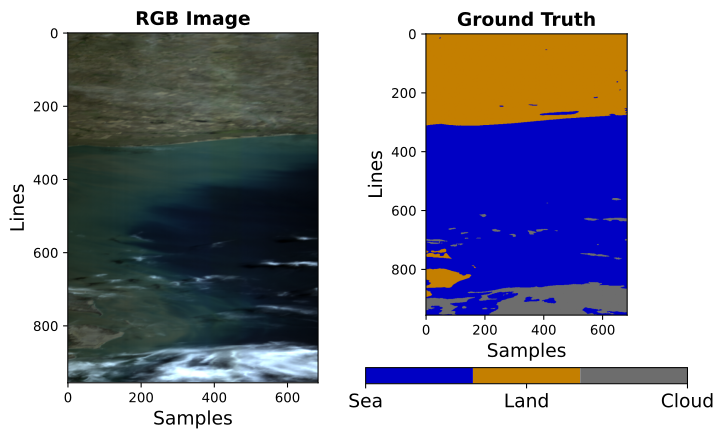}} \hspace{4mm}
\subfloat[\texttt{XGBoost}]{\includegraphics[scale=0.42]{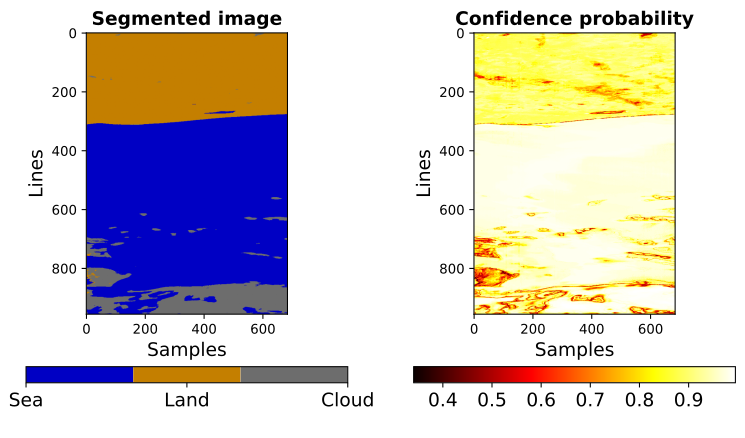}}\\
\subfloat[\texttt{1DJuLiNet}]{\includegraphics[scale=0.42]{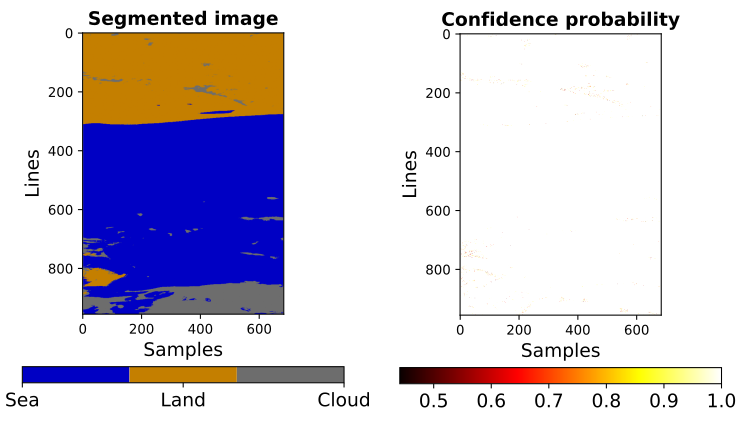}} \hspace{4mm}
\subfloat[\texttt{1DJuLiNetSingularity}]{\includegraphics[scale=0.42]{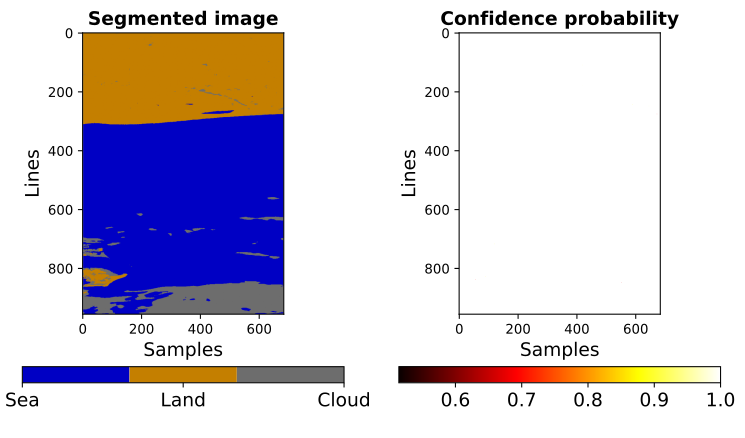}} \\
\subfloat[\texttt{1DJuLiNetSingularityF07}] {\includegraphics[scale=0.42]{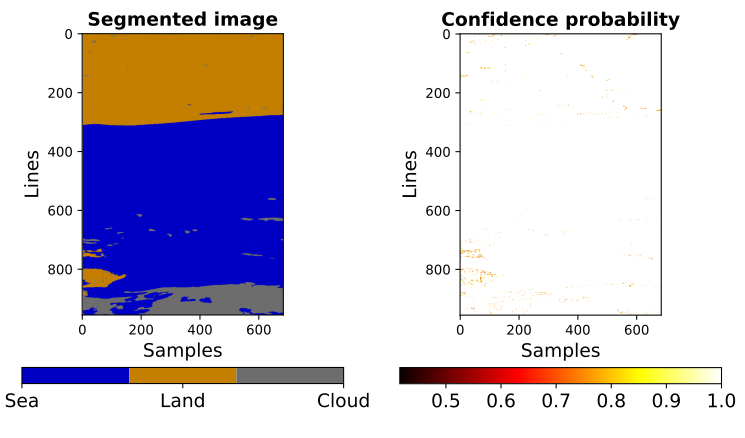}} \hspace{4mm}
\subfloat[\texttt{1DJuLiNetSingularityF30}] {\includegraphics[scale=0.42]{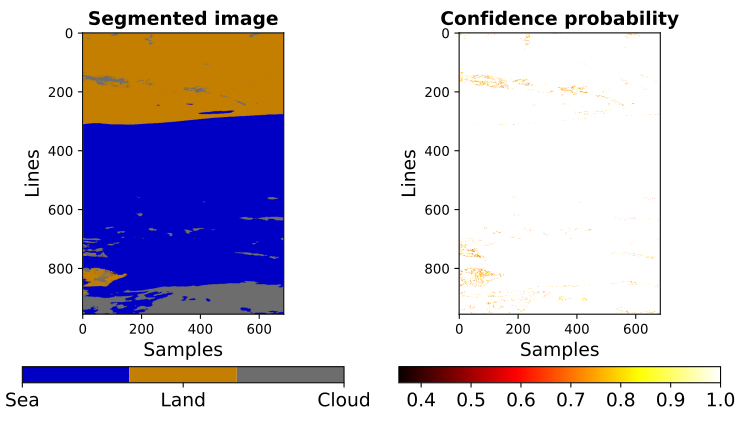}} 

\caption{Comparison of (a) Ground truth, (b) \texttt{XGBoost}, (c) \texttt{1DJuLiNet}, (d) \texttt{1DJuLiNetSingularity}, (e) \texttt{1DJuLiNetSingularityF07}, and (f) \texttt{1DJuLiNetSingularityF30} on the sample representing Argentina.}
\label{fig:sample2}
\end{figure*}

\begin{figure*}[!t]
    \centering
\subfloat[Ground Truth]{\includegraphics[scale=0.42]{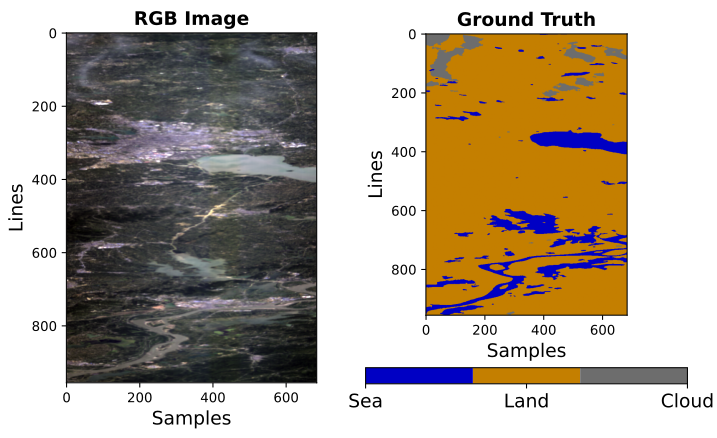}} \hspace{4mm}
\subfloat[\texttt{XGBoost}]{\includegraphics[scale=0.42]{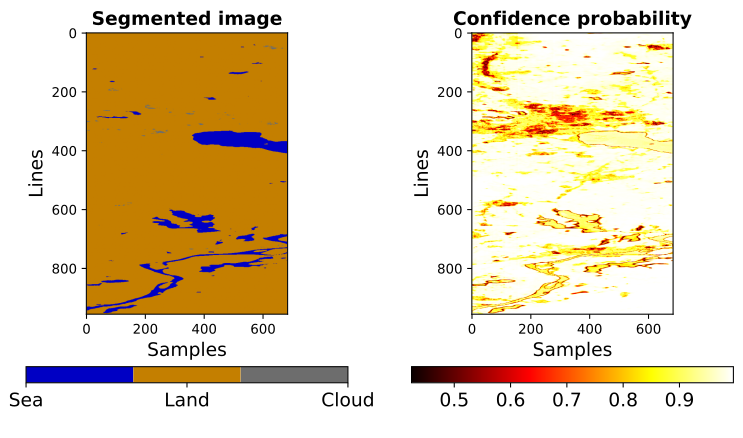}}\\
\subfloat[\texttt{1DJuLiNet}]{\includegraphics[scale=0.42]{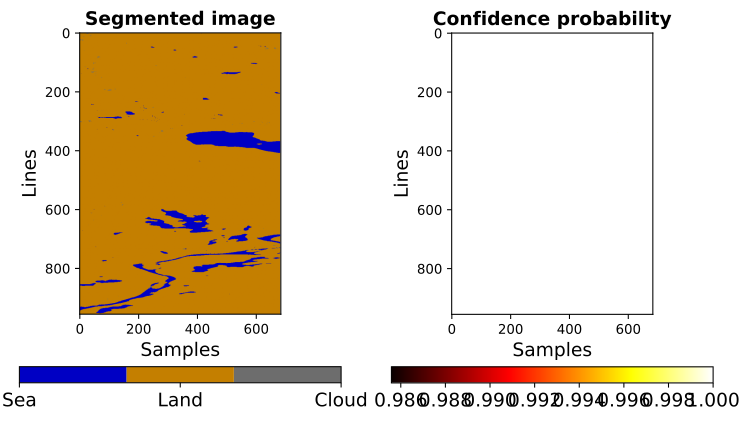}} \hspace{4mm}
\subfloat[\texttt{1DJuLiNetSingularity}]{\includegraphics[scale=0.42]{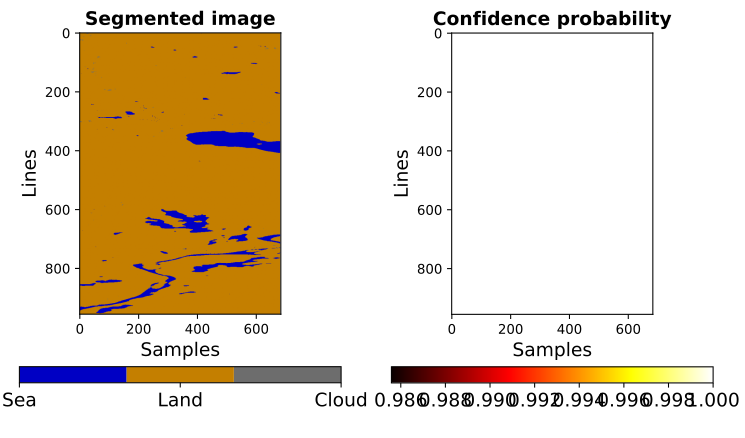}} \\
\subfloat[\texttt{1DJuLiNetSingularityF07}] {\includegraphics[scale=0.42]{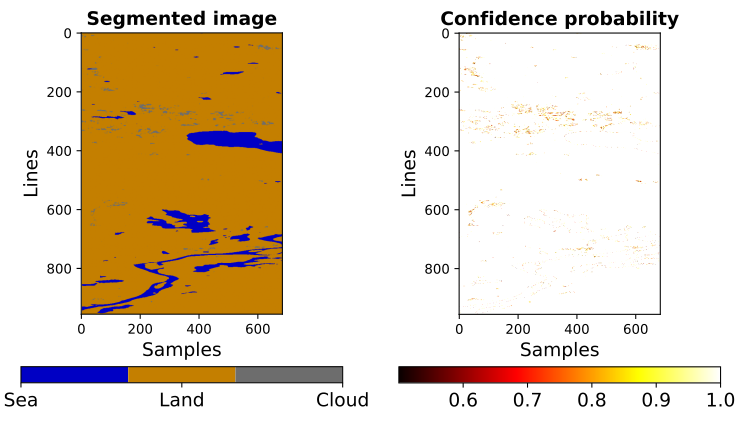}} \hspace{4mm}
\subfloat[\texttt{1DJuLiNetSingularityF30}] {\includegraphics[scale=0.42]{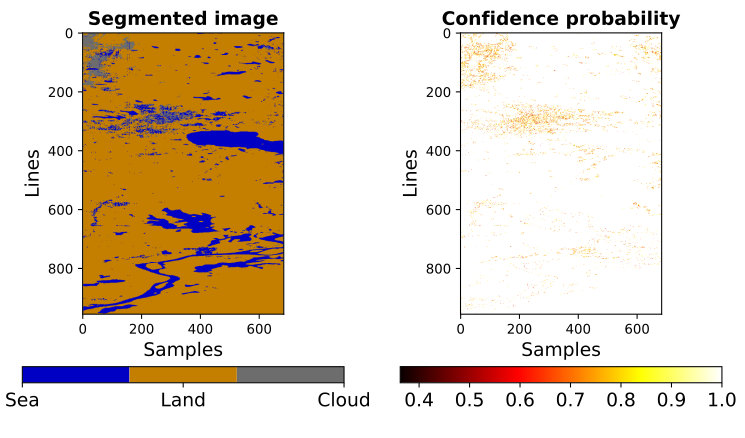}} 

\caption{Comparison of (a) Ground truth, (b) \texttt{XGBoost}, (c) \texttt{1DJuLiNet}, (d) \texttt{1DJuLiNetSingularity}, (e) \texttt{1DJuLiNetSingularityF07}, and (f) \texttt{1DJuLiNetSingularityF30} on the sample representing China.}
\label{fig:sample3}
\end{figure*}

\begin{figure*}[!t]
    \centering
\subfloat[Ground Truth]{\includegraphics[scale=0.42]{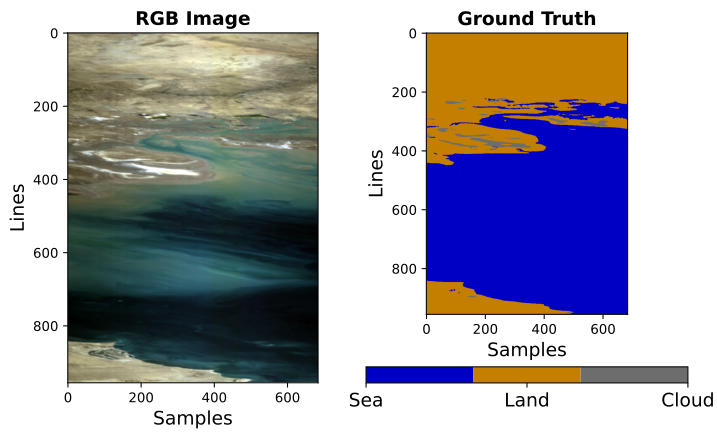}} \hspace{4mm}
\subfloat[\texttt{XGBoost}]{\includegraphics[scale=0.42]{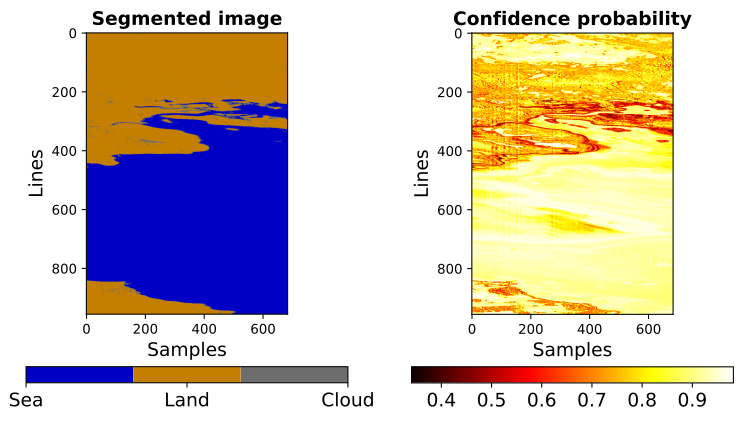}}\\
\subfloat[\texttt{1DJuLiNet}]{\includegraphics[scale=0.42]{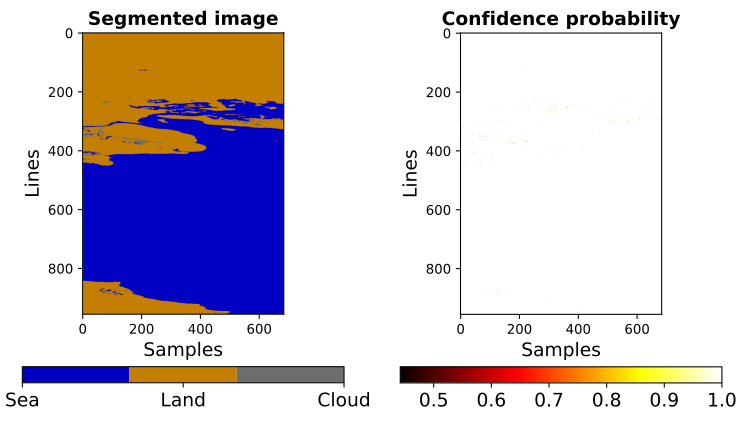}} \hspace{4mm}
\subfloat[\texttt{1DJuLiNetSingularity}]{\includegraphics[scale=0.42]{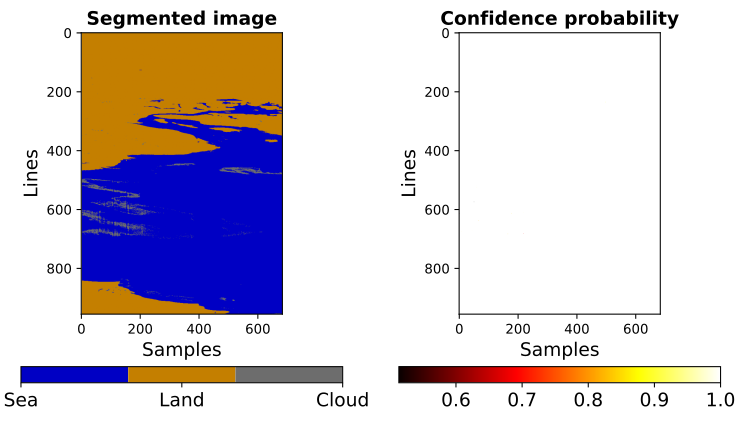}} \\
\subfloat[\texttt{1DJuLiNetSingularityF07}] {\includegraphics[scale=0.42]{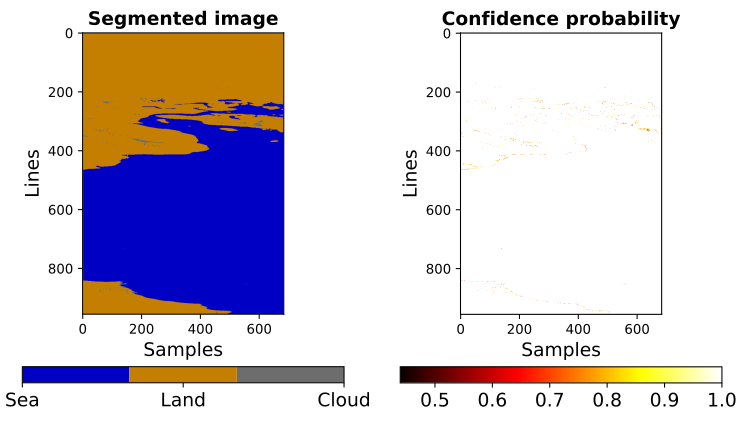}} \hspace{4mm}
\subfloat[\texttt{1DJuLiNetSingularityF30}] {\includegraphics[scale=0.42]{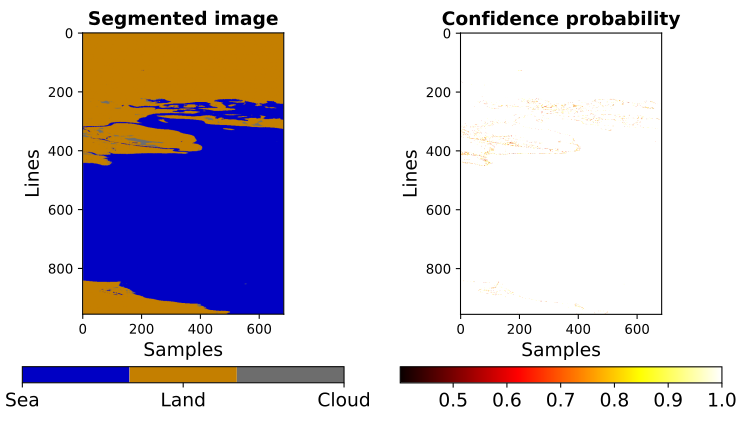}} 

\caption{Comparison of (a) Ground truth, (b) \texttt{XGBoost}, (c) \texttt{1DJuLiNet}, (d) \texttt{1DJuLiNetSingularity}, (e) \texttt{1DJuLiNetSingularityF07}, and (f) \texttt{1DJuLiNetSingularityF30} on the sample representing Iran.}
\label{fig:sample4}
\end{figure*}

\begin{figure*}[!t]
    \centering
\subfloat[Ground Truth]{\includegraphics[scale=0.42]{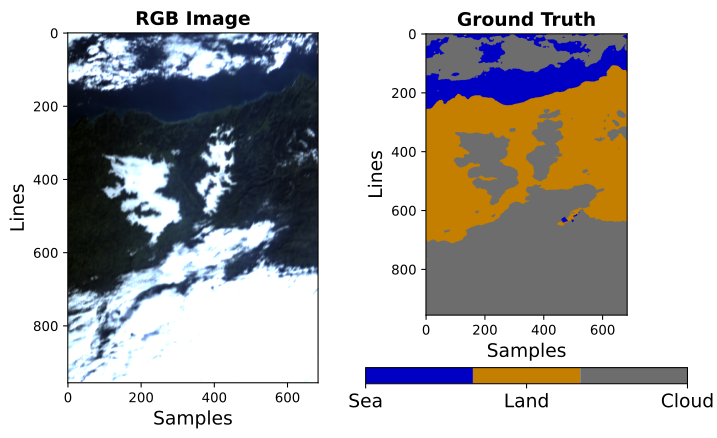}} \hspace{4mm}
\subfloat[\texttt{XGBoost}]{\includegraphics[scale=0.42]{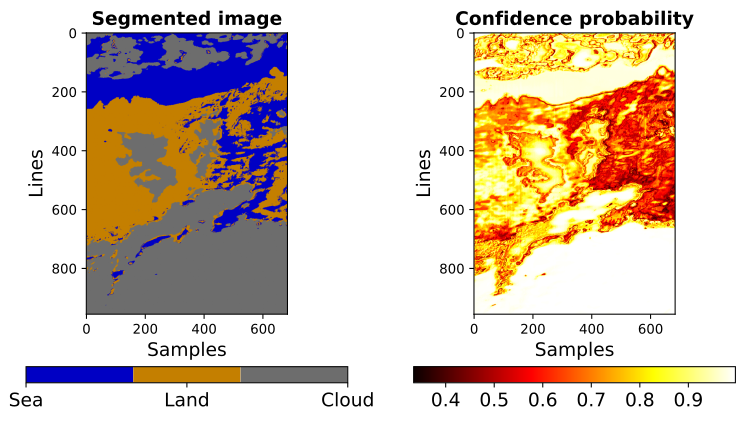}}\\
\subfloat[\texttt{1DJuLiNet}]{\includegraphics[scale=0.42]{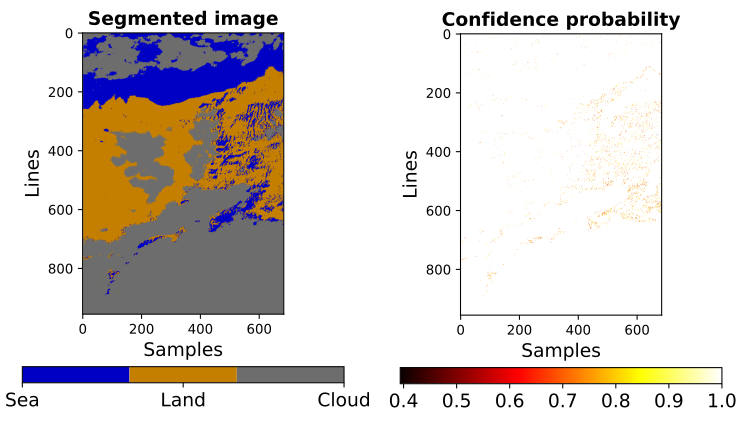}} \hspace{4mm}
\subfloat[\texttt{1DJuLiNetSingularity}]{\includegraphics[scale=0.42]{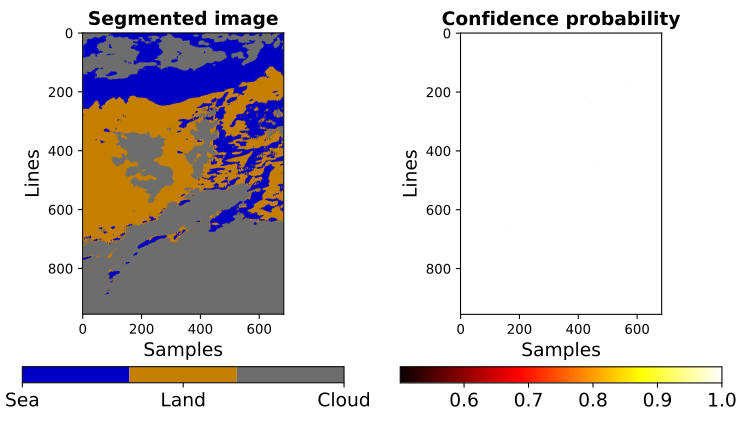}} \\
\subfloat[\texttt{1DJuLiNetSingularityF07}] {\includegraphics[scale=0.42]{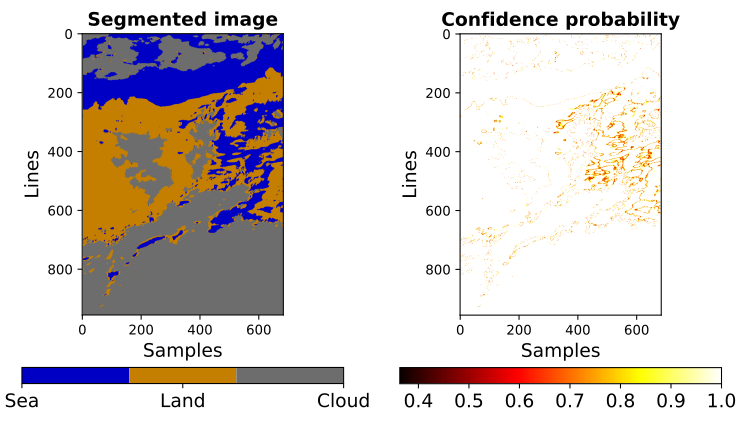}} \hspace{4mm}
\subfloat[\texttt{1DJuLiNetSingularityF30}] {\includegraphics[scale=0.42]{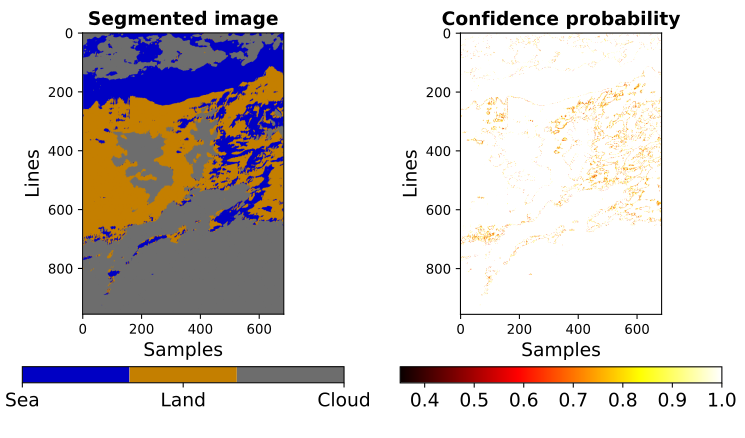}} 

\caption{Comparison of (a) Ground truth, (b) \texttt{XGBoost}, (c) \texttt{1DJuLiNet}, (d) \texttt{1DJuLiNetSingularity}, (e) \texttt{1DJuLiNetSingularityF07}, and (f) \texttt{1DJuLiNetSingularityF30} on the sample representing Spain.}
\label{fig:sample5}
\end{figure*}

\subsection{Accuracy Experiments}
To identify appropriate evaluation metrics, it is important to align them with the practical goals of the application. The segmentation model is to be executed onboard the satellite to classify each pixel as cloud, land, or water, with the aim of estimating cloud coverage prior to image transmission. Based on this estimation, the system decides whether or not to download a given image. Since the presence of clouds is the main criterion for rejecting an image, performance on the cloud class is of primary importance. In particular, precision for the cloud class is critical, as false positives -- pixels incorrectly labeled as cloud -- can lead to discarding images that contain valuable land or water information. Additionally, the Jaccard index (IoU) provides a useful measure of the spatial accuracy of cloud segmentation, capturing the degree of overlap between predicted and true cloud regions. Finally, given the relatively mild class imbalance in this dataset, the difference between overall accuracy (BOA) and balanced overall accuracy remains small.

Table \ref{tab:accuracy} shows the classification results of the aforementioned metrics, the binary metrics being displayed for the cloud class only.
As expected, the CNN model achieves the highest performance in nearly all the evaluation metrics, though the other models also demonstrate strong performance. Fig.~\ref{fig:confusion} presents the confusion matrices for each model, from which we can see that generally all the models are good predictors in all classes.
The the most populated class is ``land'' with about $\sim1.71 \times 10^6$ samples, followed by ``sea'' with $\sim1.25 \times 10^6$ samples, and finally ``cloud'' with about $\sim0.46 \times 10^6$ samples.
We observe that the \texttt{1DJuLiNetSingularityFx} models tend to be better predictors for the ``sea'' class.
While, the \texttt{1DJuLiNetSingularity}, when compared to \texttt{1DJuLiNet}  model,
mislabels significantly less true ``land'' for cloud, leaking more into the ``sea'' class.

Furthermore, Fig.~\ref{fig:sample1} illustrates the pixel-level classification of the 5 samples from the test set. Each 
figure presents the actual image followed by the predicted, the ground-
truth labeled and confidence probability. All models perform reasonably well on this test set, with a consistent misclassification of land pixels as sea pixels in the third image. Notably, the CNN model exhibits very high confidence across all images, which may indicate a suboptimal data split.

\subsection{Inference Time Experiments}
Table \ref{tab:inftime} summarizes inference times on a CPU and a GPU.
For the CPU, times were 
consistent accross different images. However,  for the GPU the first 
image took much longer than subsequent images, and, hence, we provide the two time categories separately (i.e., GPU 1st Im. and GPU Rest). Note that, for \texttt{LightGBM}, we were unable to run inference on GPUs due to
lacking software functionality.

From Table \ref{tab:inftime}, it can be observed that the CNN model is the slowest in terms of inference time whereas \texttt{XGBoost} is the fastest-performing model. On CPUs, both the CNN model and its compressed variants exhibit significantly slower inference times. On GPUs, the compressed CNN achieves a notable speedup for the first image -- processing it twice as fast as the uncompressed version -- while inference times for subsequent images remain similar across all CNN variants. The CNN model with reduced features is by far the fastest among CNN-based models when running on CPUs.

\begin{table}[!t]
    \centering
    \caption{Comparison of model inference times.
    %All experiments were performed on a \texttt{g5.2xlarge} instance; hardware specifications can be found on AWS.
    The times are given per image on the test image set.}
    \begin{tabular}{llll}
        \toprule
        \textbf{Model} & \textbf{CPU} & \textbf{GPU 1st Im.} & \textbf{GPU Rest} \\
        \midrule
        \texttt{XGBoost}  & \qty{240}{\milli\second} & \textbf{\qty{122}{\milli\second}} & \qty{50}{\milli\second} \\
        \texttt{LightGBM} & \qty{300}{\milli\second} & -- & -- \\
        \texttt{1DJuLiNet} & \qty{4.82}{\second} & \qty{761}{\milli\second} & \textbf{\qty{1}{\milli\second}} \\
        \texttt{1DJuLiNetSingularity} & \qty{5.62}{\second} & \qty{316}{\milli\second} & \textbf{\qty{1}{\milli\second}} \\
        \texttt{1DJuLiNetSingularityF30} & \qty{802}{\milli\second} & \qty{710}{\milli\second} & \textbf{\qty{1}{\milli\second}} \\
        \texttt{1DJuLiNetSingularityF18} & \qty{421}{\milli\second} & \qty{254}{\milli\second} & \textbf{\qty{1}{\milli\second}} \\
        \texttt{1DJuLiNetSingularityF07} & \qty{278}{\milli\second} & \qty{215}{\milli\second} & \textbf{\qty{1}{\milli\second}} \\
        \texttt{1DJuLiNetSingularityF04} & \textbf{\qty{218}{\milli\second}} & \qty{215}{\milli\second} & \textbf{\qty{1}{\milli\second}} \\
        \bottomrule
    \end{tabular}
    \label{tab:inftime}
\end{table}

%\section{Compatibility with Components Off the Shelf (COTS) Hardware} \label{sec:cots_compatibility}
%The evaluation of commercially available satellite-grade hardware revealed that no platform met both the technical requirements and the constraints imposed by the project budget.

\section{Discussion}\label{sec:cots_compatibility}
An analytical two-step strategy is utilized to assess compatibility with components off the shelf hardware: an algorithm complexity analysis followed by a hardware feasibility mapping.

\subsection{Complexity Analysis}
The computational complexity of the model is assessed by estimating the floating point operations per second (FLOPS) required at each processing stage. This analysis enables a detailed characterization of the model’s computational load and facilitates the screening of hardware candidates capable of supporting the required workload.
In Table \ref{tab:flops}, the outcome of this evaluation is reported. 

From Table \ref{tab:flops}, it is evident how the compression and feature-reduction layers notably reduce the complexity of the model without significantly affecting performance (see Table~\ref{tab:accuracy} for a performance summary). 

\begin{table}[!t]
    \centering
    \caption{Comparison of model complexities.
    This table clearly highlights the decrease of computational complexity of various models w.r.t. \texttt{1DJuLiNet}. Compressing the model leads to an about \qty{63}{\percent} reduction in the number operations, while feature-reduction offers a gradual trade-off between performance and efficiency. It is noteworthy that boosting models do not make use of floating point operations per se, besides weighted averaging, but rather work on Floating point Comparisons (COMP).
    }
    \label{tab:flops}
%    \resizebox{\textwidth}{!}{ 
    \begin{tabular}{@{} lccc @{}}
        \toprule
        \textbf{Model}                      & \textbf{kFLOPS / pixel}   & \textbf{COMP / pixel} \\
        \midrule
        \texttt{XGBoost}                    & $\sim$ 0.066              & $\sim$ 330            \\
        \texttt{LightGBM}                   & $\sim$ 0.069              & (214–754)             \\
        \texttt{1DJuLiNet}              & 124.42                    & --                    \\
        \texttt{1DJuLiNetSingularity}    &  45.99                    & --                    \\
        \texttt{1DJuLiNetSingularityF30}         &  15.49                    & --                    \\
        \texttt{1DJuLiNetSingularityF18}         &   6.17                    & --                    \\
        \texttt{1DJuLiNetSingularityF07}         &   1.79                    & --                    \\
        \texttt{1DJuLiNetSingularityF04}         &   1.06                    & --                    \\
        \bottomrule             
    \end{tabular}
%    }
\end{table}

Besides the evaluation of the computational operations and comparison, the CPU inference time can be used as a uniform metric to compare different models. To select the most appropriate algorithm,
we display a diagram in Fig.~\ref{fig:score-vs-time},
showing different metrics for each model and the associated inference time.
This diagram showcases that computational resources can be reduced by an order of magnitude without significantly decreasing prediction quality.

\subsection{Feasibility Mapping}
The resource requirements of the model -- specifically processing throughput, memory usage, and power consumption -- are critical factors in assessing its suitability for deployment on edge devices, particularly for onboard satellite applications.

In their work, Justo et al. \cite{justo24} evaluated the feasibility of deploying their model on the hardware platform of the HYPSO-1 satellite. This platform features a Zynq-7030 System-on-Chip (SoC), which integrates a dual-core ARM Cortex-A9 CPU and a Kintex-7 FPGA for onboard data processing. The system provides 32 KB of L1 data cache for memory access, enabling storage of up to 8K model parameters when encoded in 4-byte format. They demonstrated that the \texttt{1DJuLiNet} model, comprising only of 4563 parameters, is compatible with this hardware. Subsequently, in a later study, Justo et al. \cite{Justo25} reported a successful onboard deployment of \texttt{1DJuLiNet} on the HYPSO-1 satellite.

In contrast to \texttt{1DJuLiNet}, we introduce several alternative models that exhibit lower parameter counts (see Table \ref{tab:train-time}), comparable or improved performance (while the original model achieved an estimated accuracy of \qty{93}{\percent} in \cite{justo24}, our retraining attained over \qty{95}{\percent}, with most of the compressed and boosting models surpassing \qty{93}{\percent} accuracy -- refer to Table \ref{tab:accuracy}), and significantly reduced computational demands. Among these, \texttt{1DJuLiNetSingularity} model achieves a \qty{65}{\percent} reduction in FLOPS with very similar prediction scores with half of the memory footprint (only 12 kB instead of 24 kB) and about one third of the original trainable parameters (1419 parameters, instead of 4563). Recall that \texttt{1DJuLiNetSingularity} is obtained through a tensor network based compression \cite{Orus14}, where convolutional layers are replaced by tensorized layers using the Multiverse Computing's Singularity\texttrademark\ (Deep Learning) suite. 

\begin{figure}[!t]
    \centering
    \includegraphics[width=1.05\linewidth]{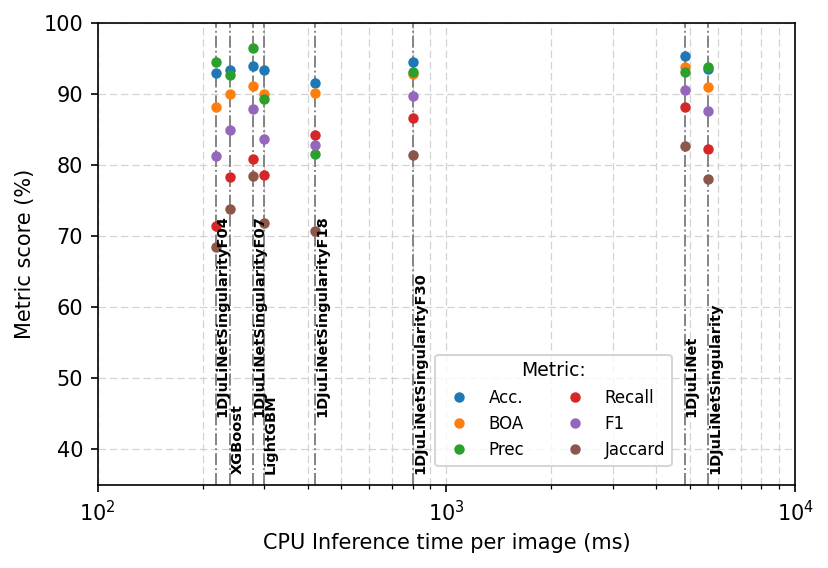}
    \caption{Diagram of prediction quality of studied models compared with their CPU's inference time per image in \qty{}{\milli\second}. Inference time, and therefore model complexity can be reduced by over one order of magnitude at the cost of a small decrease in prediction quality. }
    \label{fig:score-vs-time}
\end{figure}

Furthermore, the model \texttt{1DJuLiNetSingularityF30} reduces the input space dimension through principal component analysis achieving \qty{87.6}{\percent} FLOPs reduction, while the \texttt{XGBoost} model requires as few as \qty{66}{FLOPS} and 330 floating-point comparisons (see Table \ref{tab:flops}). These results
do not only confirm the feasibility of deploying these models onboard, but also demonstrate improvements in inference latency (see Table \ref{tab:inftime} and Figure \ref{fig:score-vs-time}), and indicate potential improvements in energy efficiency. These findings support the practical viability of deploying lightweight CNN models, particularly feature-reduced variants, in budget-constrained satellite missions.

\section{Conclusions} \label{sec:conclusions}
In this study, multiple machine learning approaches are evaluated for 
on-board cloud masking in hyperspectral satellite imagery. The investigated models include gradient boosting methods (\texttt{XGBoost} and
\texttt{LightGBM}), CNNs (\texttt{1DJuLiNet}), compressed CNNs (\texttt{1DJuLiNetSingularity}), and CNNs using feature reduction techniques (\texttt{1DJuLiNetSingularityFx}, where $\texttt{x}=4, 7, 18, 30$).  

Our experimental results demonstrate that all boosting and CNN-based 
models performed well, achieving over \qty{93}{\percent} classification 
accuracy, with CNN models exceeding \qty{95}{\percent} accuracy. 
Additionally, all models have proved to be lightweight, with trainable 
parameter counts ranging from 12 to 4500 and storage requirements 
between \qty{5}{\kilo\byte} and \qty{265}{\kilo\byte}. The 
\texttt{1DJuLiNetSingularity} model was among the smallest in storage 
at \qty{12}{\kilo\byte}, with the absolute smallest being the 
\texttt{1DJuLiNetSingularityF07} model, counting only 12 training parameters and 
requiring just \qty{5}{\kilo\byte} of storage.  

Inference speed varied by architecture and hardware:
\begin{itemize}
    \item On GPUs, all models ran efficiently, with CNNs achieving around
    \qty{1}{\milli\second} per image and \texttt{XGBoost} requiring up to \qty{50}{\milli\second} per image.  
     \item \texttt{XGBoost} presents the smallest computational footprint (with only \qty{66}{FLOPS} and 300 floating-point comparisons) with fast inference of \qty{240}{\milli\second} per image, while the most accurate CNNs were significantly slower at \qtyrange{4}{5}{\second} per image. However, the \texttt{1DJuLiNetSingularityF30} model significantly improved CPU performance, reducing inference time to
     \qty{800}{\milli\second} per image, making it a viable candidate for low-power deployment scenarios. 
\end{itemize} 

Training efficiency also varied:
\begin{itemize}
    \item Boosting models were the fastest to train, completing training in just $\sim\qty{12}{\minute}$ on both CPUs and GPUs.  
    \item CNNs required \qtyrange{27}{50}{\minute} for two epochs, while the \texttt{1DJuLiNetSingularityF30} improved training efficiency, completing in \qty{16}{\minute}.
\end{itemize}

Considering accuracy, model size, training efficiency, and inference speed,
the model \texttt{1DJuLiNetSingularityF30} emerged as the best overall choice. It maintained a high accuracy ($>\qty{94}{\percent}$), had one of the smallest storage footprints (20KB), required only 597 trainable parameters, and significantly improved inference speed on CPUs (\qty{800}{\milli\second} per image) while remaining fast on GPUs. 

This work shows that lightweight AI models can be a practical solution for real-time cloud and cloud shadow masking on hyperspectral imagery. The evaluated models performed consistently well, with high accuracy and low computational cost, even under constraints typical of satellite hardware. That said, the study focused on one dataset (HYPSO-1), which, while well-suited for benchmarking, does not cover the full range of environments or sensor configurations satellites might encounter. Moreover, although hardware compatibility was carefully assessed through
computational complexity measurements, the models were not deployed on real satellite systems in this phase. These aspects do not detract from the results but rather point to natural next steps to confirm the robustness of these models in a wider range of conditions.

\section*{Acknowledgment}
This work was supported by the ``ENFIELD: European Lighthouse to Manifest Trustworthy and Green AI'' initiative under the 1st Innovation Scheme Open Call, specifically addressing the vertical VS.1 -- ``AI Satellite On-Board Processing Model for Cloud and Cloud Shadow Masking on Hyperspectral Images with a Metadata Perspective''.

\printbibliography
\end{document}